\documentclass[10pt,twocolumn,letterpaper]{article}

\usepackage{btas}
\usepackage{times}
\usepackage{epsfig}
\usepackage{graphicx}
\usepackage{amsmath}
\usepackage{amssymb}
\usepackage{fancyhdr}

\usepackage{color}
\usepackage{caption}
\usepackage{subcaption}
\usepackage{placeins}
\usepackage{afterpage}
\usepackage{array}
\usepackage{tabularx}

\usepackage{float}
\usepackage{comment}
\usepackage{booktabs}
\usepackage{array}
\usepackage{multicol}
\usepackage[table]{xcolor}

\usepackage[pagebackref=true, breaklinks=true, letterpaper=true, colorlinks, bookmarks=false]{hyperref}

\hypersetup{
  colorlinks, linkcolor=red
}
\newcolumntype{M}[1]{>{\centering\arraybackslash}m{#1}}
\newcommand*{\mysim}{\mathord{\sim}}
\makeatletter
\g@addto@macro{\UrlBreaks}{\UrlOrds}
\makeatother

\btasfinalcopy % *** Uncomment this line for the final submission

\ifbtasfinal\fi
\begin{document}

\fancyhf{}
\fancyhead[R]{\thepage}
\fancyfoot[C]{\footnotesize 978-1-5386-7180-1/18/\$31.00 \textcopyright\ 2018 IEEE}
\renewcommand{\headrulewidth}{0pt}
\fancypagestyle{plain}{%
   \fancyhf{} 
   \fancyfoot[C]{\footnotesize 978-1-5386-7180-1/18/\$31.00 \textcopyright\ 2018 IEEE}
   \renewcommand{\headrulewidth}{0pt}
}
%

%%%%%%%%% TITLE
\title{Altered Fingerprints: Detection and Localization}

\author{Elham Tabassi, Tarang Chugh, Debayan Deb, and Anil K. Jain\\
Department of Computer Science and Engineering, Michigan State University, East Lansing, MI 48824\\
{\small E-mail:\tt{\{tabassie,chughtar,debdebay,jain\}@cse.msu.edu}}
}

\maketitle
%\thispagestyle{empty}

%%%%%%%%% ABSTRACT
\begin{abstract}

Fingerprint alteration, also referred to as obfuscation presentation attack, is to intentionally tamper or damage the real friction ridge patterns to avoid identification by an AFIS. This paper proposes a method for detection and localization of fingerprint alterations. Our main contributions are: (i) design and train CNN models on fingerprint images and minutiae-centered local patches in the image to detect and localize regions of fingerprint alterations, and (ii) train a Generative Adversarial Network (GAN) to synthesize altered fingerprints whose characteristics are similar to true altered fingerprints. A successfully trained GAN can alleviate the limited availability of altered fingerprint images for research. A database of $4,815$ altered fingerprints from 270 subjects, and an equal number of rolled fingerprint images are used to train and test our models. The proposed approach achieves a True Detection Rate (TDR) of $99.24\%$ at a False Detection Rate (FDR) of $2\%$, outperforming published results. The synthetically generated altered fingerprint dataset will be open-sourced.

\end{abstract}

%%%%%%%%% BODY TEXT
\section{Introduction}
\label{sec:intro}
%Presentation attack is a major concern for biometric deployments. In fingerprint domain, the rate of intentional alteration of fingerprints to fool fingerprint recognition algorithms is increasing. Altering a fingerprint directly changes its friction ridge pattern and avoids correct identification of the person. To mitigate this problem, presentation attack detection, i.e., methods for distinguishing live and real biometric samples from fake or spoofed samples is gaining attention and popularity. 

%The widespread deployment of Automated Fingerprint Identification Systems (AFIS) in law enforcement and border controlapplications has heightened the need for ensuring that these systems are not compromised. While several issues related to fingerprintsystem security have been investigated, including the use of fake fingerprints for masquerading identity, the problem of fingerprintalteration or obfuscation has received very little attention. Fingerprint obfuscation refers to the deliberate alteration of the fingerprintpattern by an individual for the purpose of masking his identity. Several cases of fingerprint obfuscation have been reported in thepress. 

\begin{figure}
  \includegraphics[width=\linewidth]{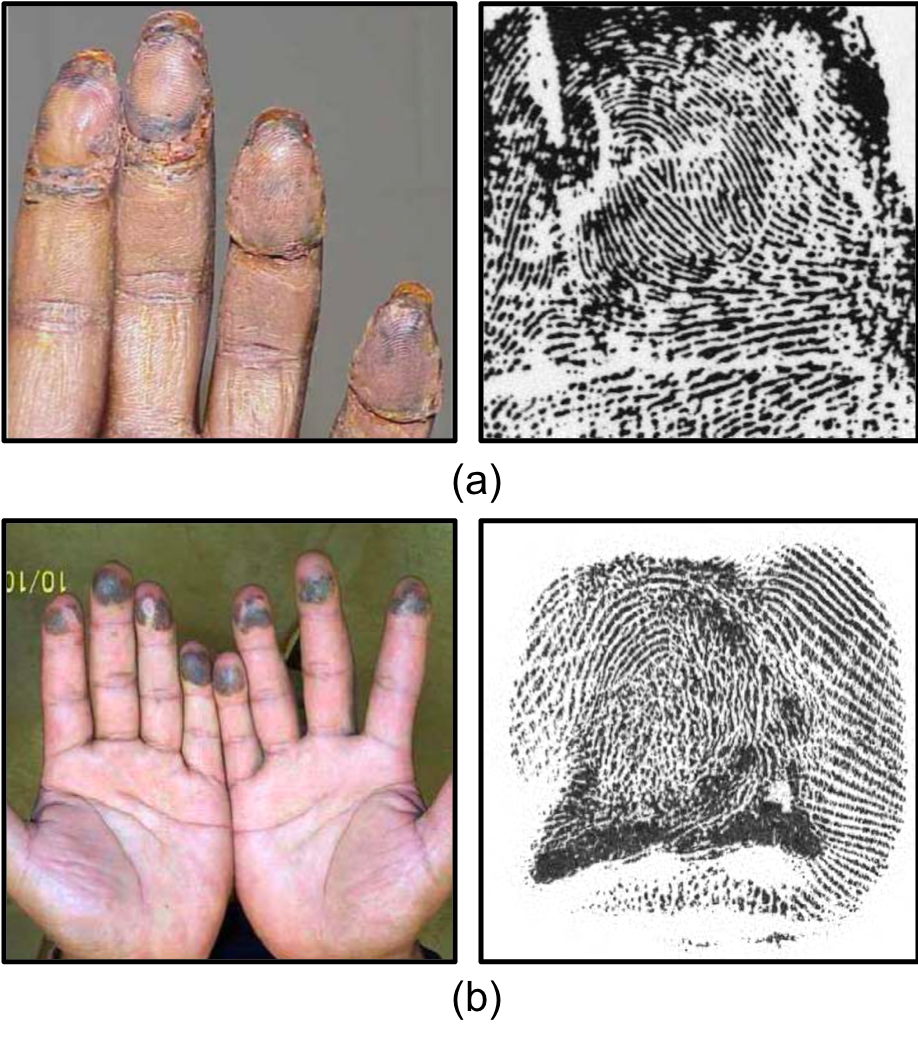}
  \caption{Example images of altered fingerprints. (a) Transplanted friction ridge skin from sole, and (b) fingers that have been bitten. Source: \cite{yoon2012altered}.}
  \label{fig:alteredfingerprints}
\end{figure}

Fingerprints have been used to identify individuals for more than a century~\cite{jain201650}. Being one of the most reliable biometrics, it has been used by law enforcement and forensic laboratories for background checks, booking suspects, and crime scene investigations. In addition, homeland security agencies deploy Automated Fingerprint Identification Systems (AFIS) for watch list comparisons of passengers arriving at ports of entry and national registry systems for de-duplication before issuing an ID. While state of the art AFIS are quite accurate and robust~\cite{fpvte}, their recognition accuracy drops when they encounter noisy fingerprint images whose friction ridges are degraded or destroyed. Intentional or unintentional destruction of friction ridge patterns, degrades the information content of a fingerprint image and therefore increases the error rate of an AFIS. Intentional fingerprint alteration, known as ``altered fingerprints'' (see Fig.~\ref{fig:alteredfingerprints}), are attempted in hope of obfuscating the true identity to evade law enforcement and is a threat to AFIS~\cite{csi}.

%\vspace*{-\baselineskip}
\begin{table*}[htp]
\footnotesize
\caption{Related and previous work on altered fingerprint detection.}
\centering
%\resizebox{\linewidth}{!}{
\begin{tabularx}{\linewidth}{M{2.6cm}M{4cm}M{5.5cm}M{3.7cm}}
\noalign{\hrule height 1.3pt}
\textbf{Source} & \textbf{Method} & \textbf{Dataset} & \textbf{Performance} \\
\noalign{\hrule height 1pt}

Feng, Jain and Ross~\cite{feng2010} & orientation field & 1,976 simulated altered fingerprints & $92\%$ detection rate at false positive rate of $7\%$ \\ \hline

Tiribuzi et al.~\cite{tiribuzi2012multiple} & minutiae density maps and orientation entropies & $1000$ genuine and synthetic altered fingerprints & $90.4\%$ classification accuracy \\ \hline
 
Yoon et al.~\cite{yoon2013there, yoon2012altered} & orientation field and minutiae distribution & $4,433$ operational altered fingerprints from 270 subjects & $70.2\%$ detection rate at false positive rate of $2.1\%$ \\ \hline
 
Ellingsgaard and Busch~\cite{handbook,IWBF} & orientation field and minutia orientation & 116 altered and 180 unaltered from various sources & $92.0\%$ detection rate at false positive rate of $2.3\%$ \\ \hline

\rowcolor{gray!50}\textbf{Proposed Approach} & input image and minutiae-based patches; CNN models & 4,815 altered and 4,815 valid fingerprints from 270 subjects & $99.24\%$ detection rate at false positive rate of $2\%$ \\
\noalign{\hrule height 1.3pt}
\end{tabularx}
\label{tab:prev}
\end{table*}%

The first recorded observation of finger tip skin transplant is by Galton who in 1896 reported his findings in a ``casual" graft of ridged skin. A man had been cutting cardboard with a sharp knife and cut his skin. A piece of skin was inadvertently sliced off. This piece was immediately applied to the wound and tightly bandaged. Examination of the injury (30 years later!) showed that the slip of skin had been successfully engrafted, though replaced at right angles to its original direction, as shown by the alignment of ridges~\cite{cummins}. Cases of tampering with fingerprints to evade detection in criminal cases were reported as early as 1935. Cummins~\cite{cummins} reported 3 cases of fingerprint alterations and presented images of before and after alterations. In 2018, Business Insider reported that like many of the FBI's most wanted criminals, Eduardo Ravelo, who was added to the FBI's 10 Most Wanted list in October 2009, was believed to have had plastic surgery to alter his fingerprints to evade authorities~\cite{businessInsider}. In recent years, border crossing applications have been targeted by altered fingerprint attacks. In 2009, ABC news reported that Japanese officials arrested ``a Chinese woman who took a particularly extreme measure to evade detection''~\cite{abcnews}. The Chinese woman had paid a plastic surgeon to swap fingerprints from her right and left hands. Patches of skin from her thumbs and index fingers were reportedly removed and then grafted onto the ends of fingers on the opposite hand. As a result, her identity was not detected when she re-entered Japan illegally. In 2014, the FBI identified 412 records in its IAFIS which indicated deliberate fingerprint alterations~\cite{forensicmag}.

Detection of altered fingerprints is of high interest to law enforcement and homeland security agencies. This paper proposes deep learning based approaches to classify input fingerprint images into two classes: valid or altered fingerprints, and to localize the regions of a fingerprint that is alterated. This can be broadened to assessing the \emph{fingerprintness} of an input image~\cite{yoon2012altered}, such that valid fingerprints, or valid region of fingerprints, shall produce high scores and altered fingerprints, or altered portion of fingerprints shall produce low or zero scores.

The organization of the paper is as follows: Previous works are outlined in Section \ref{sec:litSurvey}. Section \ref{sec:data} describes the operational valid and altered fingerprints that were used for training and testing of our network models. Models are explained in Section \ref{sec:method}, along with our synthetic altered fingerprint generation algorithm. Results are presented in Section \ref{sec:results}, and finally conclusions and future work are discussed in Section \ref{sec:conclusion}.

%-------------------------------------------------------------------------
%-------------------------------------------------------------------------
\section{Related work}
\label{sec:litSurvey}
Existing approaches for detecting fingerprint alteration have primarily explored hand crafted features to distinguish between altered and valid fingerprints. Feng et al.~\cite{feng2010} trained an SVM to detect irregularities in ridge orientation field; evaluated their method on 1,976 simulated altered fingerprints; and reported a $92\%$ detection rate at a false positive rate of $7\%$. Tiribuzi et al.~\cite{tiribuzi2012multiple} combined the minutiae density maps and the orientation entropies of the ridge-flow in order to identify the altered fingerprints. They reported a $90.4\%$ classification accuracy on a dataset of $1,000$ genuine and synthetic altered fingerprints. Yoon et al.~\cite{yoon2013there, yoon2012altered} utilized the orientation field and minutiae distribution to detect altered fingerprints. Their method was tested on a database of $4,433$ altered fingerprints from $270$ subjects, providing for $70.2\%$ correctly identified altered fingerprints at a false positive rate of $2.1\%$. Ellingsgaard and Busch in ~\cite{handbook, IWBF} discuss methods for automatically detecting altered fingerprints based on analyses of two different local characteristics of a fingerprint image: identifying irregularities in the pixel-wise orientations, and examining minutia orientations in local patches. They further suggest that alteration detection should be included into standard quality measures of fingerprints. Beyond detection of altered fingerprint, Yoon and Jain ~\cite{yoon2012} investigated feasibility of a state-of-the-art commercial fingerprint matcher to link altered fingerprints to their pre-altered mates.

Table \ref{tab:prev} summarizes previous work in altered fingerprint detection. All these methods are based on examining irregularities in orientation flow or minutia maps based on hand crafted features. Our approach differs from the previous works by using a deep learning technique to learn and evaluate salient features.

Research on altered fingerprint detection has been constrained with limited availability of data and lack of public domain altered fingerprint datasets. Furthermore, because previous works each used a different dataset, comparing their results is not feasible. To alleviate this problem, we propose a method to generate synthetic altered fingerprints.

%-------------------------------------------------------------------------
%-------------------------------------------------------------------------
\section{Altered Fingerprint Dataset}
\label{sec:data}
\begin{figure}
  \includegraphics[width=\linewidth]{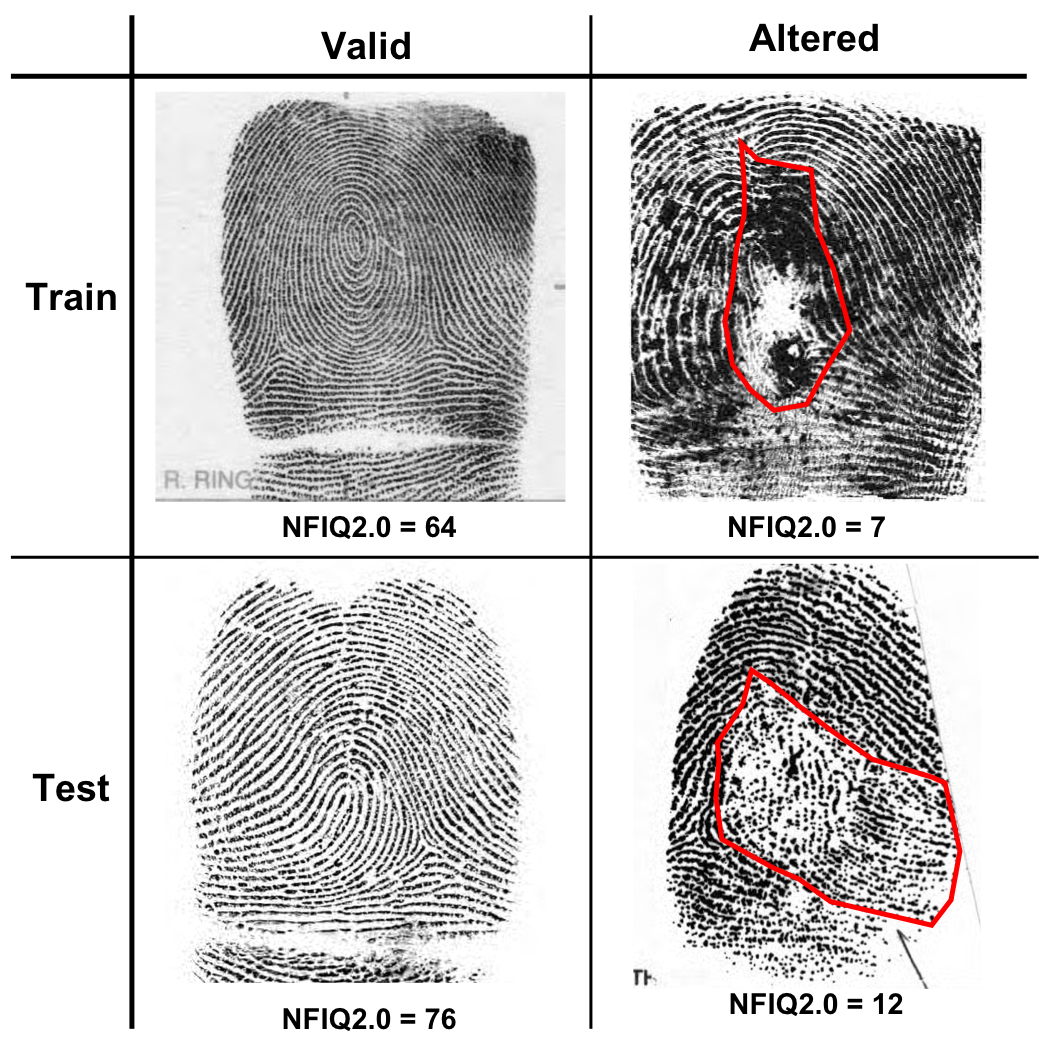}
  \caption{Example of altered and valid fingerprint images used for training and testing in one of the five folds. The altered region is highlighted in red. The NFIQ 2.0 quality scores are also presented for each image; the larger NFIQ 2.0 score, the higher fingerprint quality. The NFIQ 2.0 quality scores ranges between [0, 100].}
  \label{fig:examples}
\end{figure}

An operational dataset of $4,815$ altered fingerprints, from $635$ tenprint cards of $270$ subjects, acquired from law enforcement agencies is utilized in this study. The number of tenprint cards per subject varies from 1 to 16 due to multiple encounters. However, not all 10 fingerprint images in a tenprint card may be altered. The number of altered fingerprint instances per subject varies from 1 to 137. Another operational dataset of $4,815$ rolled fingerprint images is used for valid fingerprints. Fingerprint images in both sets of altered and valid are images collected as part of law enforcement operations. All images are 8-bits gray scale. A five-fold cross-validation is employed where in each of the five folds, the training set contains $3,852$ altered and $3,852$ valid fingerprints. The testing set in each fold contains the remaining $963$ altered and $963$ valid fingerprints, such that the train and test sets are disjoint. Figure~\ref{fig:examples} shows sample altered and valid images used for training and testing in one of the five folds.

\begin{figure}[!htb]
\begin{center}
\includegraphics[width=\linewidth]{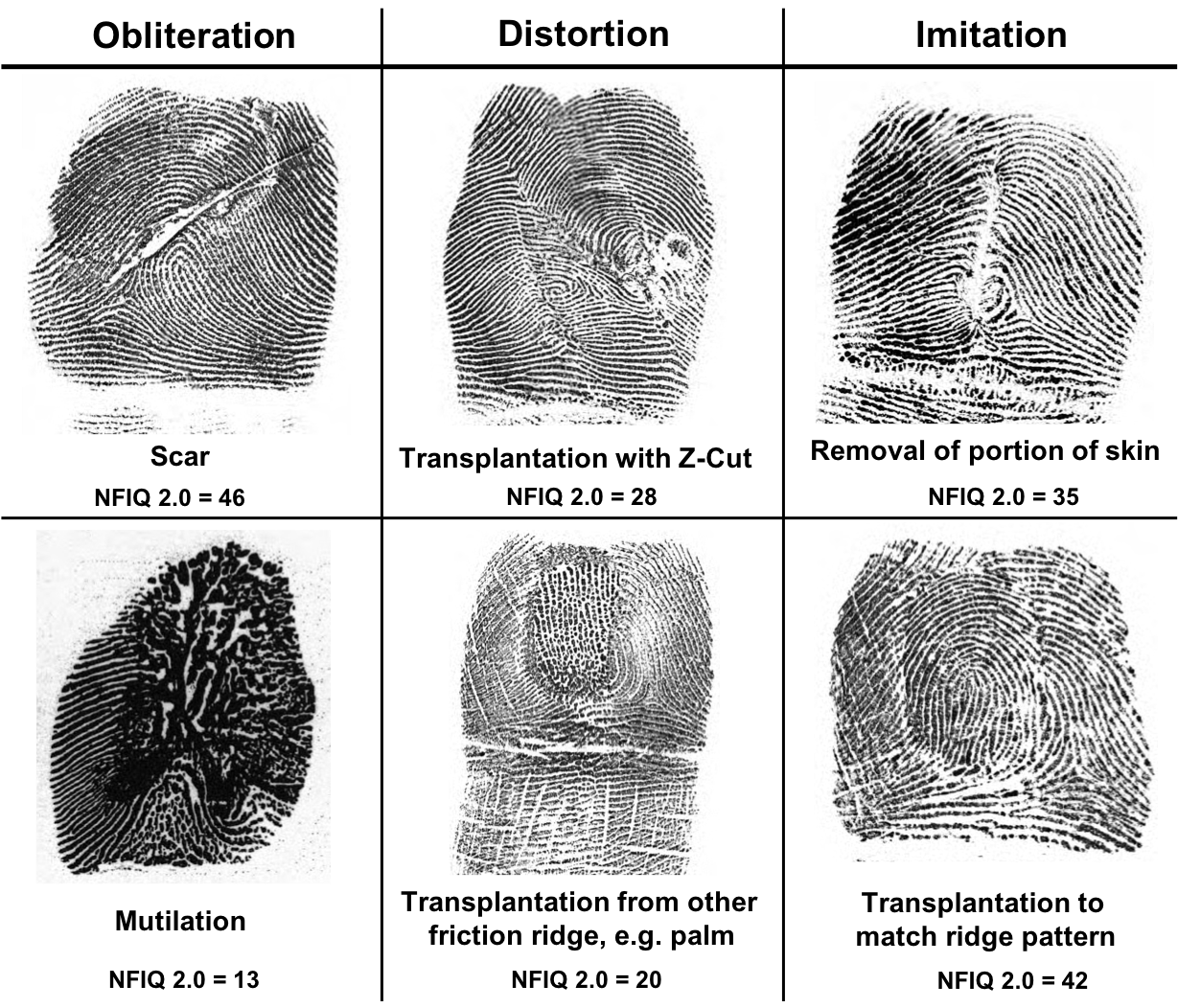}
\end{center}
   \caption{Types of fingerprint alterations: (i)~Obliteration, such as scars, or mutilations, (ii)~Distortion, \textit{i.e.} friction ridge transplantation to distort friction ridge area, and (iii)~Imitation, \textit{i.e.} transplantation or removal of friction ridge skin while still preserving fingerprint like pattern.}
\label{fig:altered}
\end{figure}

%-------------------------------------------------------------------------
%-------------------------------------------------------------------------
\section{Proposed Approach}
\label{sec:method}
 \subsection{Altered Fingerprint Detection}
 \label{sec:model}
The goal of this study is to detect altered fingerprint images. This can be formulated as a binary classification problem with two classes; \textit{altered} and \textit{valid}. A more sophisticated model would be a multi-class classification that detects the type of alteration, when valid fingerprint has a type ``none''. As shown in Figure \ref{fig:altered}, different types of alteration procedures would result in different fingerprint degradation. Different types of alteration procedures and their effect on friction ridge pattern are discussed in~\cite{yoon2012altered} and~\cite{handbook}. Based on the changes made to friction ridge patterns, they categorized altered fingerprints into three types: \textit{obliteration}, \textit{distortion}, and \textit{imitation}. 

\textit{Obliteration} consists of abrading, cutting, burning, applying strong chemicals, or transplanting friction ridge skin. Skin disease or side effects of drugs can also obliterate fingertips. \textit{Distortion} comprises of cases of using plastic surgery to convert normal friction ridge pattern into unusual ridge pattern. Some portions of skin are removed from the finger and grafted back onto a different position causing an unusual pattern. \textit{Imitation} is when surgical procedure is performed in such a way that the altered fingerprints appear as natural fingerprints, for example, by grafting skin from the other hand or a toe onto a large or perhaps the entire finger tip skin such that fingerprint ridge pattern is still preserved. While there are distinct alteration types, and despite Yoon and Jain's \cite{yoon2012altered} suggestion to develop different models for different alteration types, we propose to utilize a single model for the following two reasons: a) insufficient data for each alteration type for training deep networks, and b) manual labeling of the alteration type would be subjective because an image can suffer from more than one alteration type. We trained a Convolutional Neural Network (CNN) to classify an input fingerprint image into one of the two classes of valid or altered. Data augmentation techniques, such as mirroring, random cropping, and rotation have been employed to increase the size of the training data.

\begin{figure}[!t]
  \includegraphics[width=\linewidth]{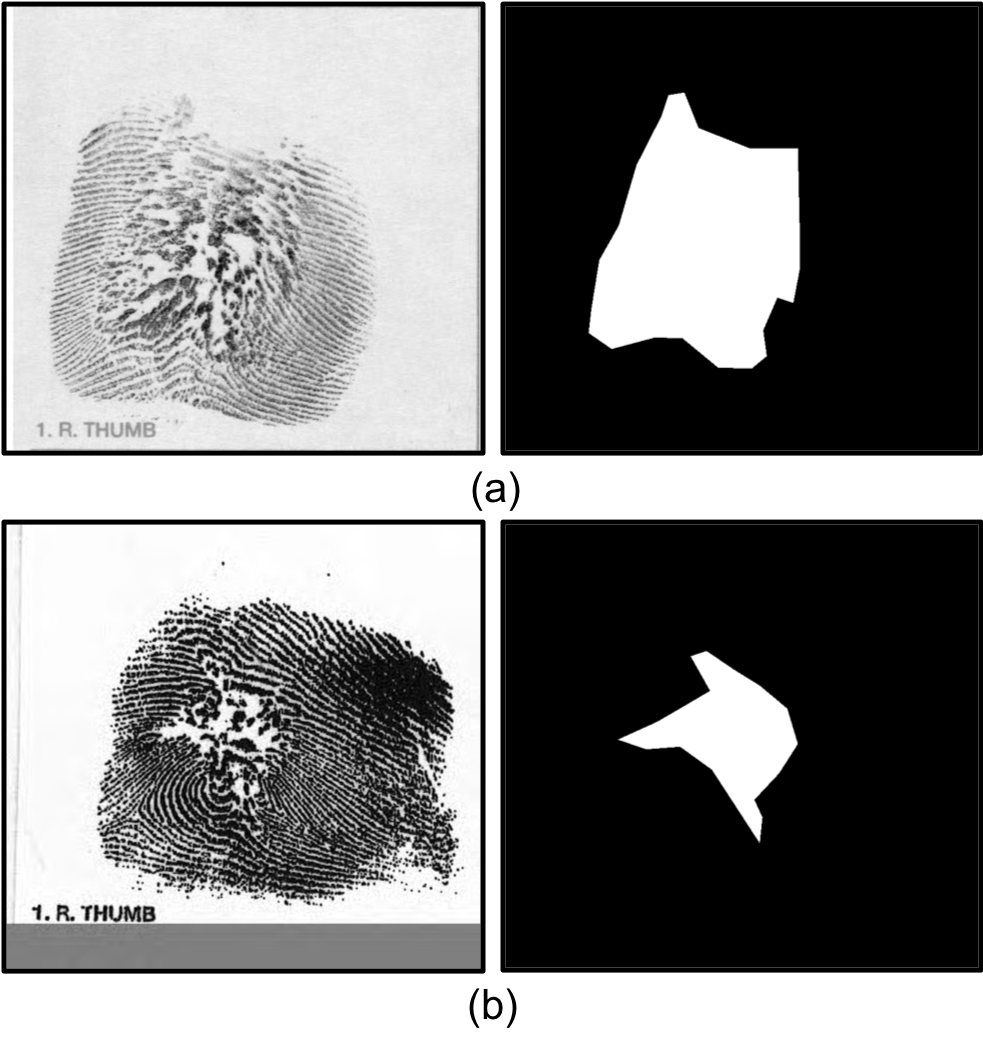}
  \caption{Examples of altered fingerprints and corresponding manually marked regions of interest (ROI) circumscribing the areas of fingerprint alterations. Local patches overlapping with manually marked ROI are labelled as altered patches, while the rest are labelled as valid. The test phase is fully automatic and does not require any manual markup.}
  \label{fig:roi}
\end{figure}

\begin{figure}[!t]
  \includegraphics[width=\linewidth]{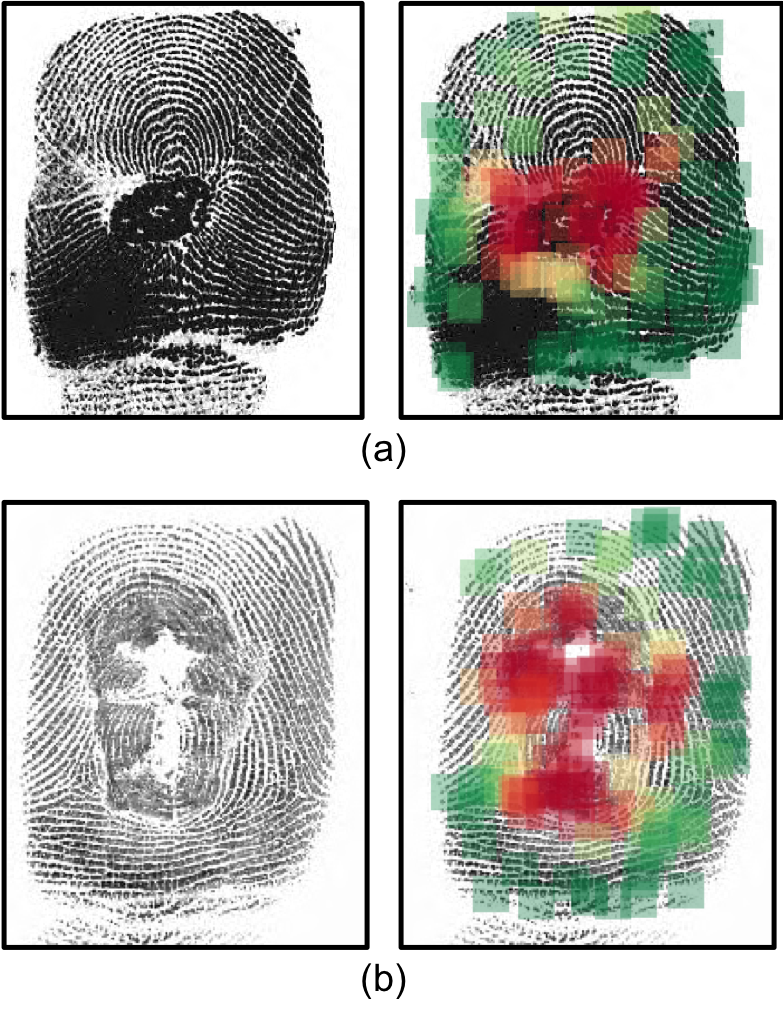}
  \caption{Examples of altered fingerprint localization by our proposed method. Local regions highlighted in red represent the altered portion of the fingerprint, whereas regions highlighted in green reflect the valid friction ridge area.}
  \label{fig:localization}
\end{figure}

%
%-------------------------------------------------------------------------
\subsection{Altered Fingerprint Localization}

\begin{figure*}
  \includegraphics[width=\textwidth]{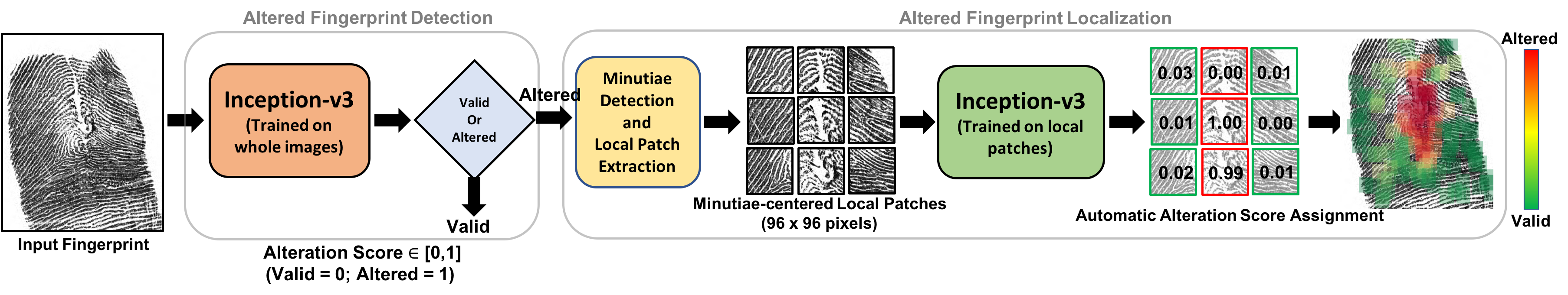}
  \caption{An overview of the proposed approach for detection and localization of altered fingerprints. We trained two convolutional neural networks (Inception-v3 and Mobilenet-v1) using full fingerprint images and local patches of images where patches are centered on minutiae locations.}
  \label{fig:flowchart}
\end{figure*}

To localize and highlight the altered regions of fingerprints, we augment our whole image based altered fingerprint detection with a patch-based approach. Our approach is as follows: First, region of interest (ROI) is manually marked for $1,182$ randomly selected altered fingerprints from our database of $4,815$ altered fingerprints. See Figure~\ref{fig:roi}. Next, local patches of size $96 \times 96$ centered around each extracted minutia are cropped. Local patches with more than $50\%$ overlap with the manually marked ROI are labelled as altered patches, and the remaining patches are labelled as valid. Because a majority of fingerprint alterations generate discontinuities and noisy regions in the friction ridge pattern, a much higher number of spurious minutiae are generated in altered fingerprints compared to valid fingerprints of the same size~\cite{yoon2012altered}. Local patches centered around minutiae have also shown to provide superior performance in fingerprint spoof detection compared to patches extracted in a raster scan manner~\cite{chugh2018fingerprint}. A total of $81,969$ valid and $89,979$ altered patches are extracted and utilized for training two different networks: Inception-v3~\cite{szegedy2016rethinking} and MobileNet-v1~\cite{mobilenet}. Fig.~\ref{fig:localization} presents examples of altered fingerprint localization output by the proposed approach. An overview of the proposed approach to detect and localize altered fingerprints is presented in Figure~\ref{fig:flowchart}. 

\begin{figure}[!t]
\begin{center}
   \includegraphics[width=\linewidth,trim={3cm 29cm 6cm 31cm},clip]{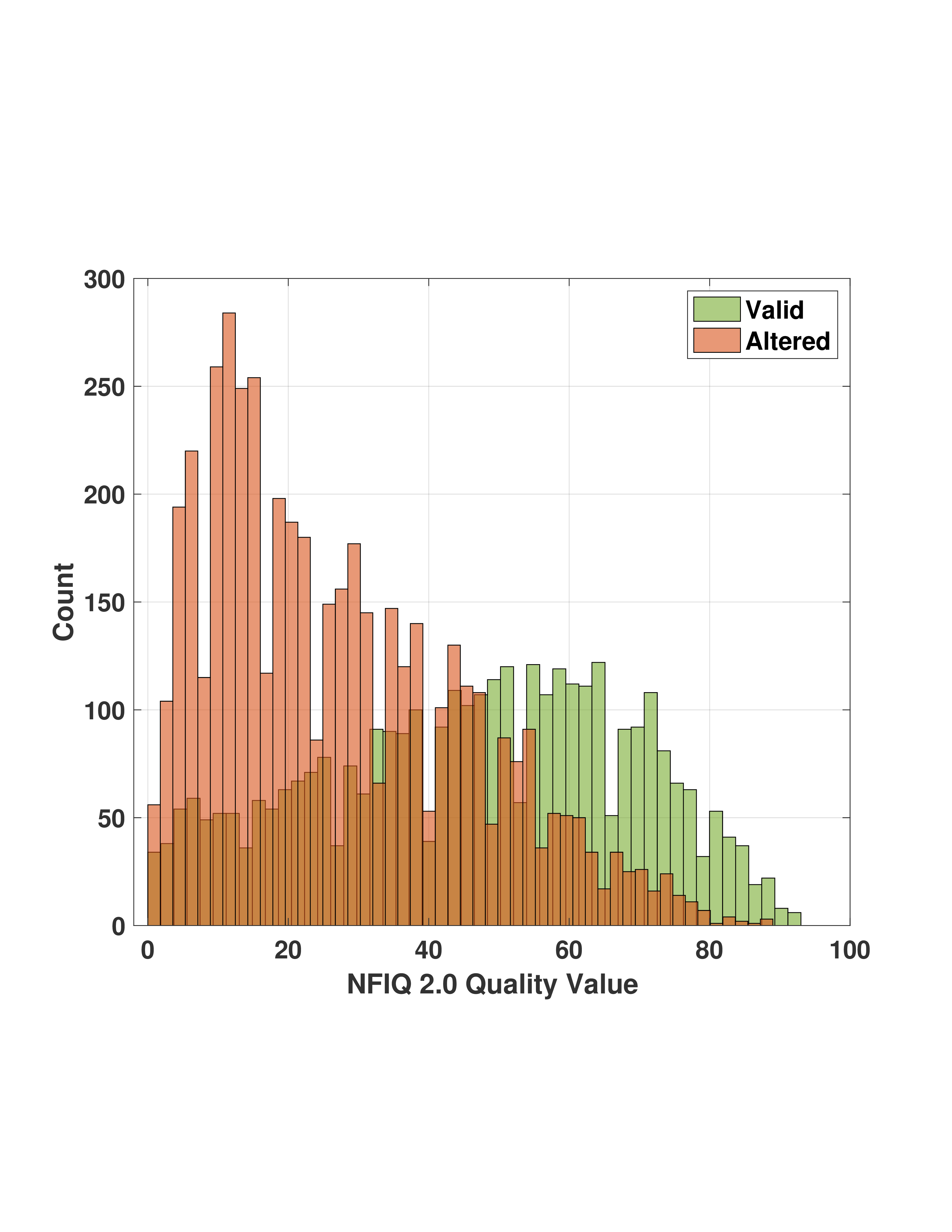}
\end{center}
   \caption{Histogram of NFIQ 2.0 quality scores for valid (green) and altered (red) fingerprint images. Approximately, 75\% of altered fingerprint images have a NFIQ 2.0 score of 40 or lower, and only 10\% of altered dataset has a NFIQ 2.0 score of larger than 50. The median NFIQ 2.0 score for altered fingerprint images is 23, while median NFIQ 2.0 score for valid fingerprint images is 48. This suggests NFIQ 2.0's suitability for detecting altered fingerprints, particularly for cases of fingerprint obliteration.} 
\label{fig:nfiq2}
\end{figure}

\subsection{Fingerprint image quality analysis}
Figure \ref{fig:nfiq2} shows distribution of NFIQ 2.0 ~\cite{nfiq2, iso} scores for the altered and valid fingerprint images used in this study. NFIQ 2.0 software reads a fingerprint image, computes a set of quality features from the image, and uses these features to predict the utility of the image as an integer score between 0 and 100. About 75\% of altered fingerprints have a NFIQ 2.0 score of 40 or lower, and only 10\% of images have a NFIQ 2.0 score larger than 50. The median NFIQ 2.0 score is 23 for altered fingerprints, and 48 for valid. This suggests that NFIQ 2.0 may be suited to detect altered fingerprints. 

%-------------------------------------------------------------------------
\subsection{Deep learning for detecting altered fingerprints}
\label{sec:learning}

Using the code in \cite{githubcode}, we were able to train MobileNet-v1~\cite{mobilenet} and Inception-v3~\cite{szegedy2016rethinking} networks %\footnote{Our implementation is available on github at [url removed for blind review].} 
as binary classifiers (altered vs. valid fingerprints). The input is a full fingerprint image and the output is a probability (or score) of belonging to Altered or Valid class, referred to as \textit{alteration score}. A valid fingerprint image should result in an alteration score of close to 0, whereas an altered fingerprint image should result in an alteration score of close to 1. The network hyper-parameters used to train the CNN models are presented in Table~\ref{tab:hyperparamters}.

%\vspace*{-\baselineskip}
\begin{table}[!t]
\footnotesize
\caption{Network hyper-paramters utilized in training CNN and GAN models.}
\centering
\begin{tabularx}{\linewidth}{M{1.8cm}M{1.9cm}M{1.9cm}M{1.25cm}}
\noalign{\hrule height 1.3pt}
\textbf{Hyper-paramters} & \textbf{Inception-v3} & \textbf{MobileNet-v1} & \textbf{DC-GAN} \\
\noalign{\hrule height 1pt}

\textbf{Batch Size} & 32 & 100 & 64 \\ \hline

\textbf{Optimizer} & RMSProp & RMSProp & Adam \\ \hline
 
\textbf{Learning Rate}  & [0.01 - 0.0001]; exp. decay 0.94 & [0.01 - 0.0001]; exp. decay 0.94 & 0.0002 \\ \hline
 
\textbf{Momentum} & 0.9 & 0.9 & 0.5 \\ \hline

\textbf{Iterations} & 25,000 & 25,000 & 1,350 \\

\noalign{\hrule height 1.3pt}
\end{tabularx}
%}
\label{tab:hyperparamters}
\end{table}%

%\begin{comment}
%\begin{figure}[htbp]
%\begin{center}
%   \includegraphics[width=\linewidth]{./figures/mobilenet.png}
%\end{center}
%   \caption{Architecture of MobileNet. Source: Howard, et al. ``MobileNets: Efficient Convolutional Neural Networks for Mobile Vision Applications'' \cite{mobilenet} }
%\label{fig:mobilenet}
%\end{figure}
%\end{comment}

 %https://github.com/ronny3050/AlteredFingerprintDetection.

\begin{figure*}[!htb]
\begin{center}
   \includegraphics[width=\linewidth]{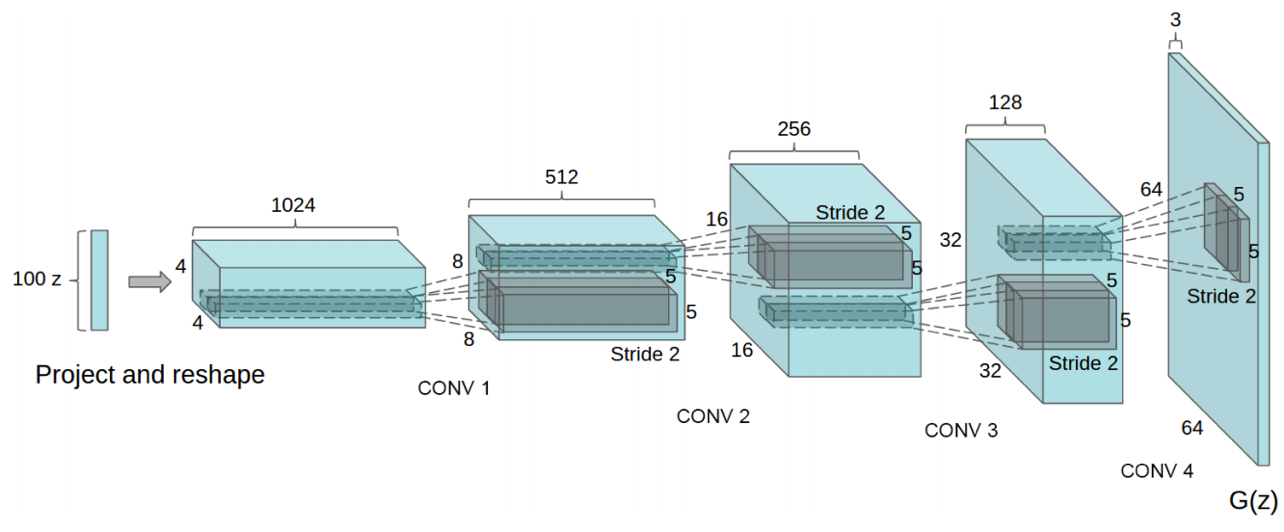}
\end{center}
   \caption{Architecture of DC-GAN used to generate synthetic altered fingerprint images. Source: Radford, et al.~\cite{gan}. }
\label{fig:gan}
\end{figure*}

%-------------------------------------------------------------------------
\subsection{Generating synthetic altered fingerprints}
\label{sec:gan}

One major constraint of studies on altered fingerprint detection is the limited amount of altered fingerprint images available. This limitation is perhaps the cause of relatively few investigations on this topic. To remedy the issue of limited available data, we trained a Generative Adversarial Networks (GAN) that generates altered fingerprints. We used the DC-GAN architecture\footnote{\url{https://github.com/carpedm20/DCGAN-tensorflow}} proposed in~\cite{gan}. See Figure~\ref{fig:gan}. We utilized all of the $4,815$ altered fingerprint images for training by cropping them to $512 \times 512$ pixels. The trained model outputs $256 \times 256$ synthetic altered fingerprint images\footnote{To avoid the fast convergence of the discriminator network, the generator network is updated twice for each discriminator network update}. %$^,$\footnote{Our implementation is available on github at [url removed for blind review].}. 
The network hyper-parameters used to train the GAN model are presented in Table~\ref{tab:hyperparamters}.

Generation of synthetic altered fingerprint is a related but separate topic than detection of altered fingerprint as discussed in Section \ref{sec:learning}. Our motivation was to investigate ways to remedy the lack of publicly available altered fingerprint data for research.
%
%
%Our implementation will be available at [url removed for blind review].%https://github.com/ronny3050/AlteredFingerprintGeneration.
%
%
%-------------------------------------------------------------------------
%-------------------------------------------------------------------------

\begin{figure}[htbp]
\begin{center}
   \includegraphics[width=\linewidth]{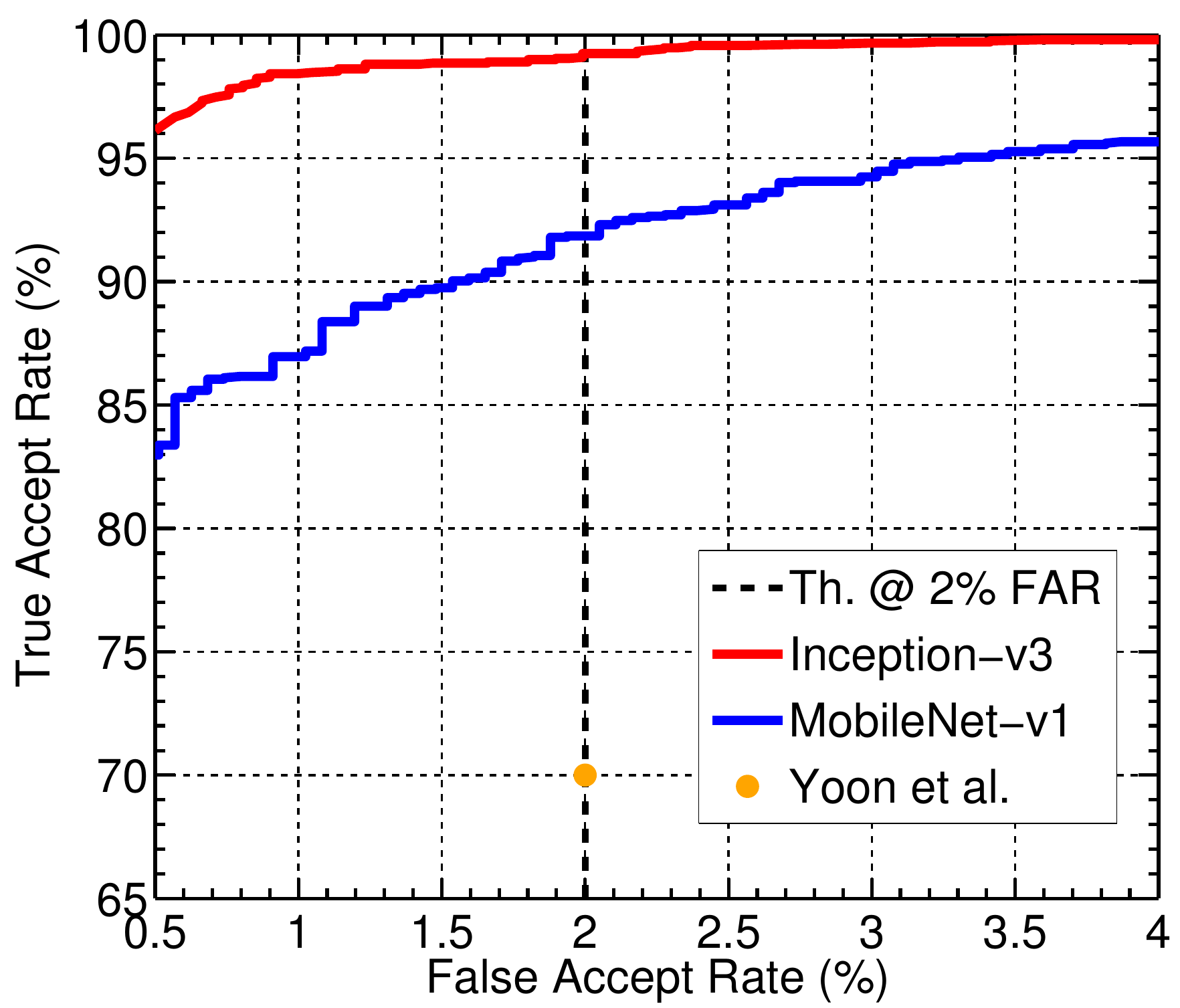}
\end{center}
   \caption{Performance curves for the proposed altered fingerprint detection approach utilizing Inception-v3 and MobileNet-v1 CNN models. Yoon et al.~\cite{yoon2012altered} (baseline) achieved a TDR of 70\% $@$ FDR = 2\% on 4,433 altered fingerprints, while the proposed approach achieves a TDR (over five folds) of 99.24\% $\pm$ 0.58\% $@$ FDR = 2\% on 4,815 altered fingerprints.}
   % Training set contains 1,756 altered and 1,756 non-altered (or valid) fingerprint images. Similarly, the test set contains 1,756 altered and 1,756 non-altered (or valid) images. All images are operational images. Altered fingerprint images are from Department of Homeland Security and the FBI. Valid (non-altered) fingerprint images are from the Michigan State Police. }
\label{fig:roc1}
\end{figure}

\begin{figure}[!t]
\centering
\includegraphics[width=\linewidth,trim={1.8cm 0cm 2.5cm 1cm},clip]{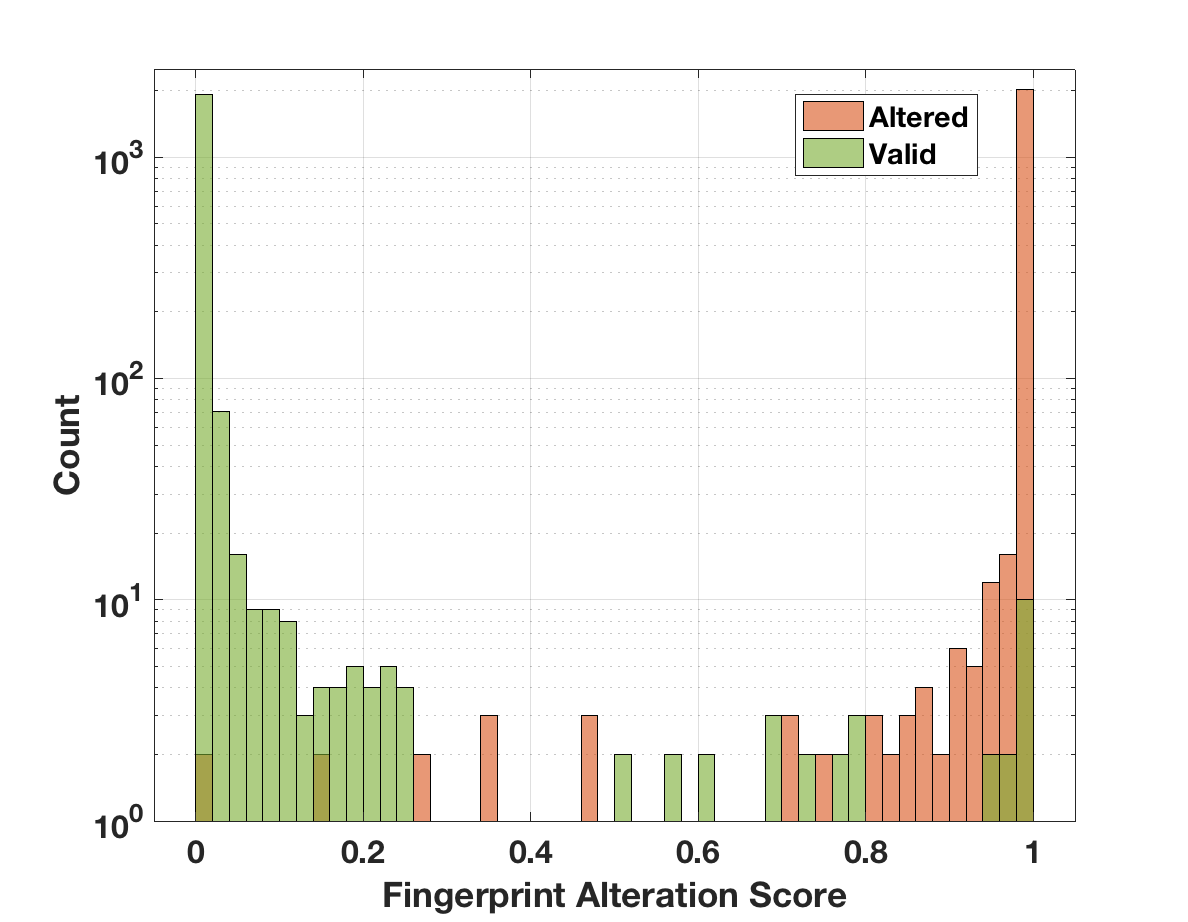}
   \caption{Alteration score histograms for valid and altered fingerprints obtained by the proposed approach using the best performing Inception-v3 model. The small overlap between the valid and altered score distributions is an indication of high discrimination power of the model. Note that the Y-axis is presented in log scale.}
\label{fig:boxplots}
\end{figure}

\begin{figure}[!t]
\centering
\includegraphics[width=\linewidth]{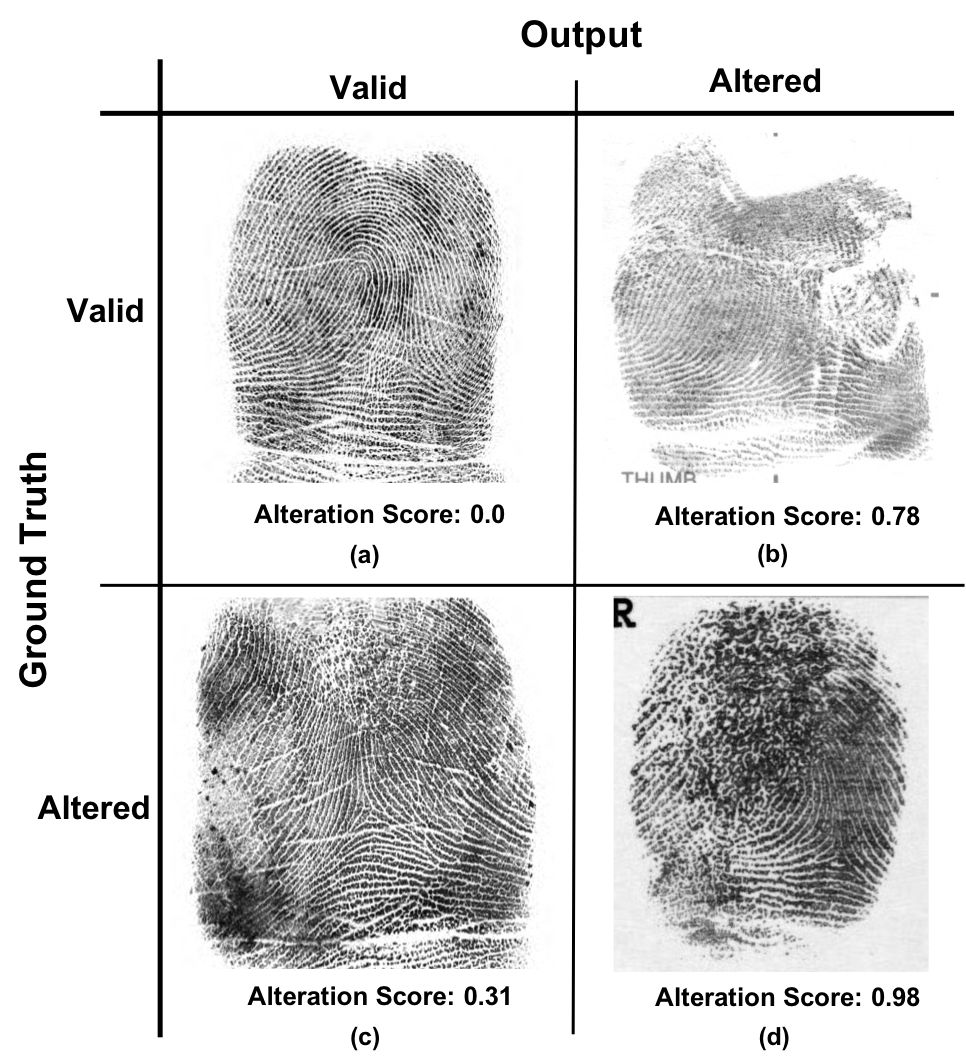}
   \caption{Example classifications and their alteration scores output by the proposed approach. (a) and (d) present correctly classified images, while (b) and (c) present incorrect classifications. (b) a valid fingerprint that receives a high alteration score primarily due to the noisy region on the right. (c) contains a small region of alteration which is similar to the noise present in valid fingerprints.}
\label{fig:classification}
\end{figure}

\section{Experimental Results}
\label{sec:results}
\subsection{Altered fingerprint detection and localization}
\label{sec:learningResult}
\begin{figure}[!t]
  \centering
    \includegraphics[width=\linewidth]{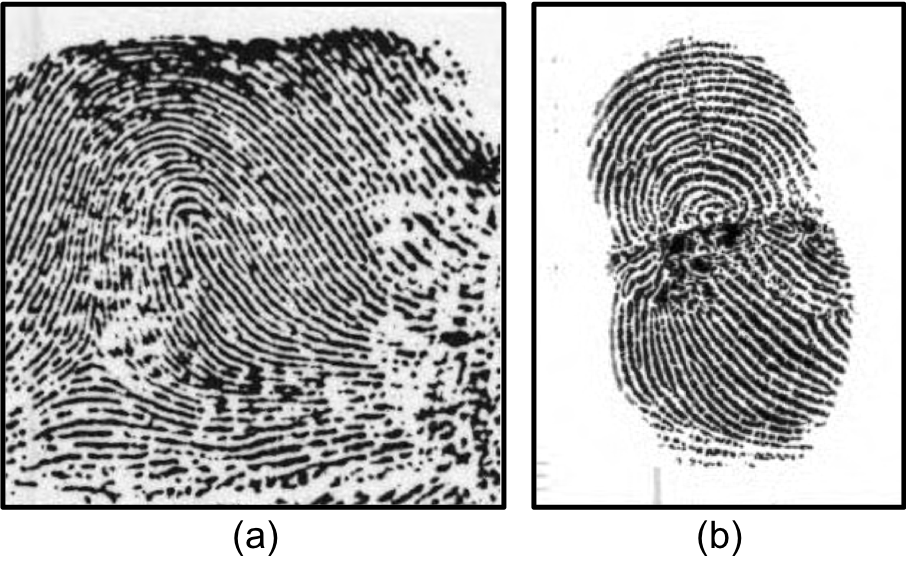}
  \caption{Example images with possible groundtruth labeling error. (a) Incorrectly labeled as altered, and (b) incorrectly labelled as valid. The Inception-v3 model outputs an alteration score of 0.20 and 0.97 for (a) and (b), respectively, indicating (a) as valid and (b) as altered.}
  \label{fig:gtError}
\end{figure}

Figure \ref{fig:roc1} shows the Receiver Operating Characteristic (ROC) curves for the proposed altered fingerprint detection approach (Inception-v3 and MobileNet-v1) compared with state-of-the-art~\cite{yoon2012altered}. The red curve shows the accuracy of the Inception-v3 implementation and the blue curve shows the accuracy of the MobileNet-v1 implementation. Inception-v3 outperforms MobileNet-v1 architecture ($\mysim99\%$ to $\mysim92\%$), while the computational requirement\footnote{We utilized NVIDIA GTX 1080 Ti GPU to run our implementation of Inception-v3 and MobileNet-V1 based altered fingerprint detection.} for MobileNet-v1 (6 ms) is almost 10 times lower compared to time required by the Inception-v3 architecture (50 ms). The superior performance of Inception-v3 over Mobilenet-v1 network can be attributed to (i) deeper convolutional network providing higher discrimination power, and (ii) larger input image size; $299 \times 299$ for Inception-v3, compared to $224 \times 224$ for Mobilenet-v1. Both network models show better detection performance than Yoon and Jain \cite{yoon2012altered} which had a true detection rate of only $70.2\%$ at a false positive rate of $2\%$. Figure \ref{fig:boxplots} shows the histograms of scores produced by our Inception-v3 model for valid and altered fingerprint images. The very small overlap of the two distributions is an indication of the high accuracy of our model. We further investigated the images that were incorrectly labeled by our model according to the ground truth labels given at the time of training. Our visual inspection of these images suggests that some of images labeled as valid, look like altered fingerprints. This might be either due to intentional alteration or cases of poor quality where fingerprint characteristics are degraded because of age or occupation (bricklayers, for example, are known to have poor quality fingerprints because their skin is severely damaged). On the other hand, some of the images labeled as altered, have a relatively small portion of the image as altered and most parts of the image look valid. If our model classifies these so called altered fingerprints as valid fingerprint images, it may not be far from truth. Example images of correct and incorrect classification by the Inception-v3 model are shown in Figure \ref{fig:classification} along with the scores generated by our model. Examples of incorrect groundtruth label are shown in Figure \ref{fig:gtError}.

To evaluate the localization of fingerprint alterations, a two-fold cross validation is performed. Two Inception-v3 networks are trained using $81,969$ valid and $89,979$ altered patches, achieving an average EER of $8.5\%$. 

\begin{figure*}[!htb]
\begin{center}
   \includegraphics[width=\linewidth]{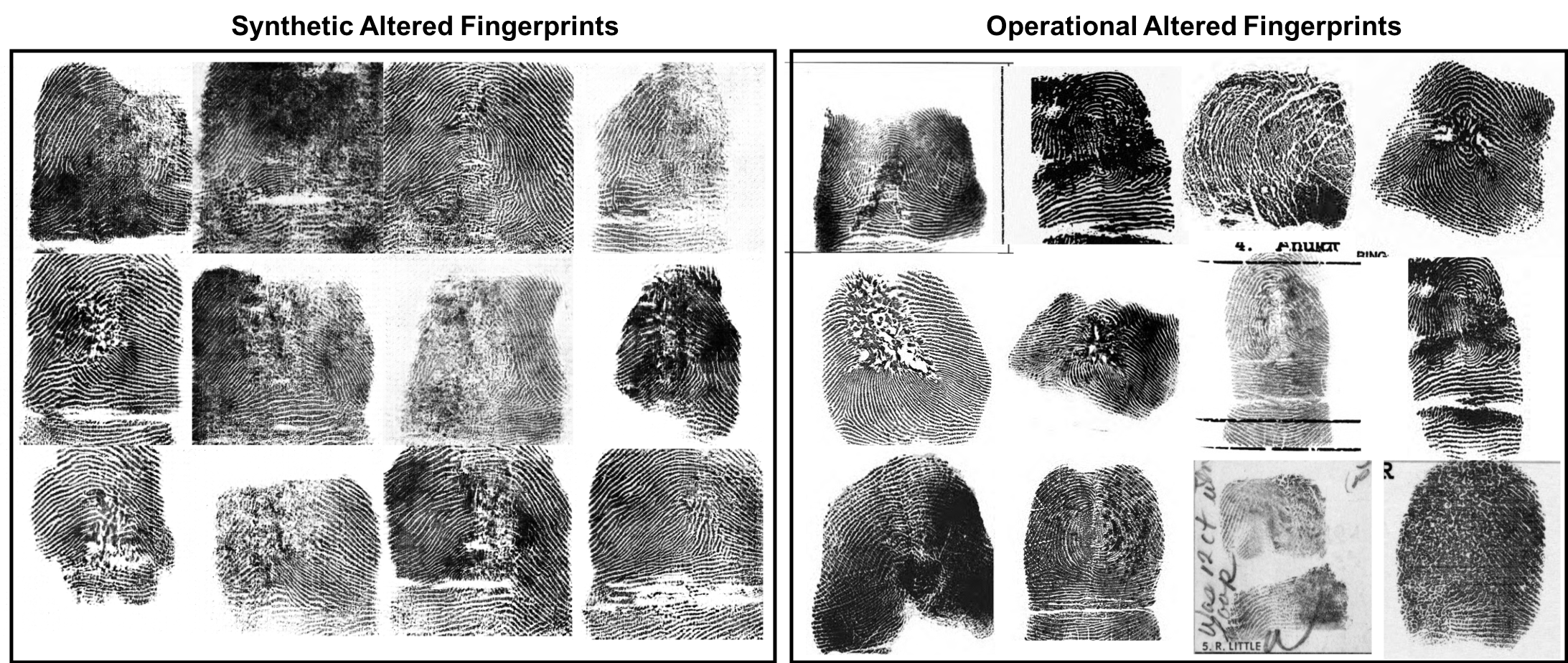}
\end{center}
   \caption{Example images of synthetic altered fingerprint images generated by the proposed GAN, compared to the operational altered fingerprint images.}
\label{fig:synthetic}
\end{figure*}

%\begin{figure}[htb]
%\centering
%\begin{subfigure}[b]{0.32\linewidth}
%  \centering
%    \includegraphics[width=\linewidth]{figures/samples_GAN/generated_1.png}
%    \caption{NFIQ2.0 = 30}
%  \end{subfigure}\hfil
%\begin{subfigure}[b]{0.32\linewidth}
%  \centering
%    \includegraphics[width=\linewidth]{figures/samples_GAN/generated2.png}
%    \caption{NFIQ2.0 = 20}
%  \end{subfigure}\vfil
%\begin{subfigure}[b]{0.32\linewidth}
%  \centering
%    \includegraphics[width=\linewidth]{figures/samples_GAN/generated3.png}
%    \caption{NFIQ2.0 = 14}
%  \end{subfigure}\hfil
%\begin{subfigure}[b]{0.32\linewidth}
%  \centering
%    \includegraphics[width=\linewidth]{figures/samples_GAN/generated4.png}
%    \caption{NFIQ2.0 = 0}
%  \end{subfigure}
%  \caption{Examples of synthetic altered fingerprint images generated using the proposed GAN}
%\label{fig:synthetic}
%\end{figure}
%-------------------------------------------------------------------------
\subsection{Generating synthetic altered fingerprints}
\label{sec:ganResults}
\begin{figure}[!t]
  \centering
    \includegraphics[width=\linewidth,trim={1.5cm 0.4cm 2.5cm 2cm},clip]{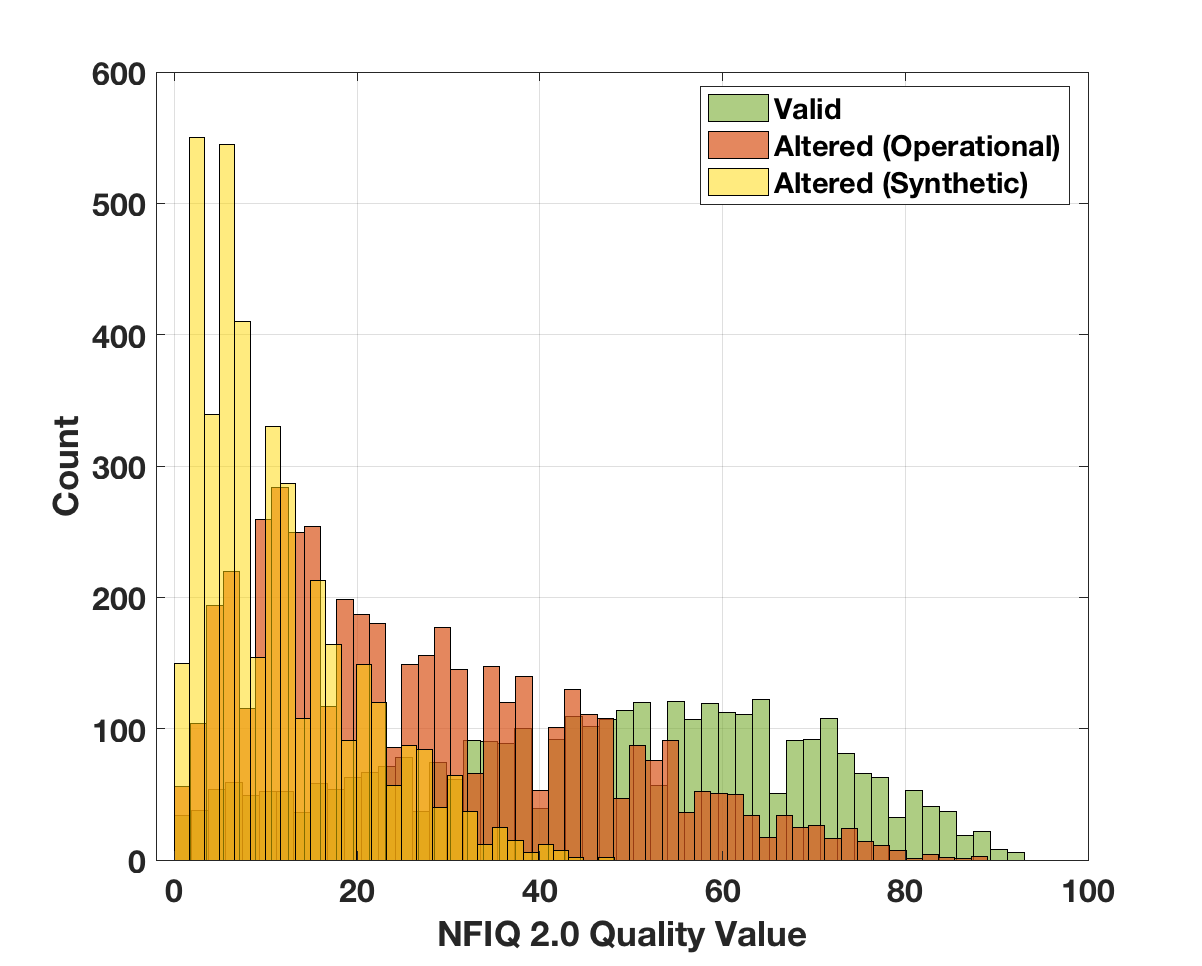}
  \caption{NFIQ 2.0 quality score distributions for 4,060 synthetically generated altered (yellow), 4,815 altered (red), and 4,815 valid fingerprint images (green). The mean NFIQ 2.0 quality scores for synthetic altered, operational altered, and operational valid fingerprint images are 11, 27, and 46, respectively.}
  \label{fig:gannfiq2}
\end{figure}

A total of $4,060$ synthetic altered fingerprint images are generated using the GAN model discussed in Section~\ref{sec:gan}. Figure~\ref{fig:synthetic} presents example images of synthetically generated altered fingerprints, compared with operational fingerprint images. The distribution of NFIQ 2.0 quality scores for synthetically generated altered, operational altered, and valid fingerprint images are shown in Figure~\ref{fig:gannfiq2}. The NFIQ 2.0 score distribution of synthetically generated altered fingerprints have large overlap with the distribution of operational altered fingerprints. The mean NFIQ 2.0 quality score for synthetic altered, altered, and valid fingerprint images are 11, 27, and 46, respectively. The low NFIQ 2.0 quality scores for synthetic altered fingerprint images can be attributed to the noisy friction ridge structure mimicked by the GAN, as well as the low resolution of the GAN output. As the training dataset is limited, a lower resolution for synthetic altered fingerprints is selected to avoid over-fitting. The synthetic altered fingerprint images are $256 \times 256$ pixels, while the operational altered fingerprint images are $750 \times 800$ pixels. This generator is our first attempt at solving the limited availability of altered fingerprint datasets, and requires further refining to match the characteristics between generated and true altered fingerprints, which we will pursue as a future work.

%To examine whether the synthetically generated fingerprint images contain similar characteristics as the operational dataset, we trained an Inception-v3 CNN model on $4,060$ synthetic altered fingerprint images and $3,852$ operational live images, and evaluated the model using the remaining 963 operational altered and the remaining 963 operational live images\footnote{The 4,815 operational live images were divided into two disjoint sets of 3,852 images for training, and 963 images for testing.}. The model was trained with the same network hyper-parameters as discussed earlier in Section \ref{sec:model} (See Table \ref{tab:hyperparamters}). Figure~\ref{fig:roc_synth} presents the ROC curve for the trained model. The assumption is that if the accuracy of the two Inception-v3 networks (one trained using operational altered fingerprint images and the other trained using synthetic fingerprint images, both using the same valid fingerprint images and the same network architecture and hyper-parameters) are similar, then the synthetic and operational altered fingerprints have similar friction ridge characteristics.

%

%
%-------------------------------------------------------------------------
%-------------------------------------------------------------------------
\section{Conclusions and Future work}
\label{sec:conclusion}

A robust and accurate method for altered fingerprint detection is critical to ensure the security of widely deployed AFIS in a variety of government and commercial applications. In this study, we have trained a CNN model using operational datasets of $4,815$ altered and $4,815$ valid fingerprint images for altered fingerprint detection. Additionally, we trained another model using minutia-centered local patches to automatically localize the regions of fingerprint alterations. Our altered fingerprint detection model achieves a True Detection Rate (TDR) of $99.24\%$ $@$ False Detection Rate (FDR) of $1\%$, compared to the previous stat-of-the-art result of TDR = 70\% at FDR = $2\%$ which used a smaller operational dataset. Finally, we trained a GAN, using the operational altered fingerprint database, to generate synthetic altered fingerprint images with similar characteristics as that of operational database. The synthetically generated altered fingerprint database will be open-sourced to alleviate the limited availability of altered fingerprint database and encourage further research on this topic.

In future, we plan to perform an analysis of pre- and post-altered fingerprint images of the same finger to benchmark the effect of alteration on recognition accuracy. We will also refine the GAN network to improve the characteristics of synthetic altered fingerprints, control the type of alteration, and use fingerprint comparison scores to assess the goodness of fit for our proposed synthetic altered fingerprint generation model.

%------------------------------------------------------------------------
{\small
\bibliographystyle{ieeetr}
\bibliography{AltFp_BTAS2018_CameraReady}
}

\end{document}